\documentclass[conference, camera]{IEEEtran}
\IEEEoverridecommandlockouts

\usepackage{cite}
\usepackage{amsmath,amssymb,amsfonts}
\usepackage{algorithmic}
\usepackage{graphicx}
\usepackage{textcomp}
\usepackage{booktabs}
\usepackage{xcolor}
\def\BibTeX{{\rm B\kern-.05em{\sc i\kern-.025em b}\kern-.08em
    T\kern-.1667em\lower.7ex\hbox{E}\kern-.125emX}}
\begin{document}
\IEEEoverridecommandlockouts
% \IEEEpubid{\makebox[\columnwidth]{ 979-8-3503-7608-1/24\$31.00 \copyright2024 IEEE \hfill} \hspace{\columnsep}\makebox[\columnwidth]{ }}

\title{\textsc{CircuitSynth}: Leveraging Large Language Models for Circuit Topology Synthesis \\
}
% \author{\IEEEauthorblockN{Anonymous Authors}}
\author{
\IEEEauthorblockN{ Prashanth Vijayaraghavan$^*$}
\IEEEauthorblockA{
\textit{IBM Research}\\
San Jose, CA 95120\\
prashanthv@ibm.com}
\and
\IEEEauthorblockN{Luyao Shi$^*$}
\IEEEauthorblockA{
\textit{IBM Research}\\
San Jose, CA 95120 \\
luyao.shi@ibm.com}
\and 
\IEEEauthorblockN{Ehsan Degan}
\IEEEauthorblockA{
\textit{IBM Research}\\
San Jose, CA 95120\\
edehgha@us.ibm.com
}
\and
\IEEEauthorblockN{Xin Zhang}
\IEEEauthorblockA{
\textit{IBM Research}\\
Yorktown Heights, NY 10598 \\
xzhang@us.ibm.com
}

}

\maketitle
% \IEEEpeerreviewmaketitle
\def\thefootnote{*}\footnotetext{These authors contributed equally to this work}\def\thefootnote{\arabic{footnote}}
\begin{abstract}
% Circuit topology generation plays a crucial role in the design of electronic circuits, influencing factors such as performance, power consumption, and overall efficiency. With the exponential growth and power of large language models (LLMs), there emerges a promising opportunity to leverage their capabilities for automatic generation of circuit topologies. In this paper, we introduce \textsc{CircuitSynth}, a novel approach that harnesses LLMs to facilitate the automated synthesis of valid circuit topologies. To this end, we curate a dataset comprising both valid and invalid circuit configurations, utilizing the capabilities of an open-source electronic circuit simulator (SPICE). Leveraging this dataset, \textsc{CircuitSynth} employs a sophisticated methodology, encompassing three key steps: (a) training a circuit validity classifier to evaluate the probability of a circuit's validity; (b) fine-tuning an LLM to generate circuit topologies; and (c) enhancing the model by refining its outputs with the aid of a circuit validity classifier. Experimental results demonstrate the effectiveness of our \textsc{CircuitSynth} model compared to various LLM variants tuned for generating valid circuit topologies. Our approach lays the foundation for future research directions aimed at enhancing efficiency and specifying output voltage, thus enabling the automated generation of circuit topologies with improved performance and adherence to design requirements.

Circuit topology generation plays a crucial role in the design of electronic circuits, influencing the fundamental functionality of the circuit. In this paper, we introduce \textsc{CircuitSynth}, a novel approach that harnesses LLMs to facilitate the automated synthesis of valid circuit topologies. With a dataset comprising both valid and invalid circuit configurations, \textsc{CircuitSynth} employs a sophisticated two-phase methodology, comprising Circuit Topology Generation and Circuit Topology Refinement. Experimental results demonstrate the effectiveness of \textsc{CircuitSynth} compared to various fine-tuned LLM variants. Our approach lays the foundation for future research aimed at enhancing circuit efficiency and specifying output voltage, thus enabling the automated generation of circuit topologies with improved performance and adherence to design requirements.

% \footnote{\textcolor{red}{Verify: Is paper anonymous-- Used IEEEpeerreviewmaketitle; Alternate Title- CircuitLLM/TopoSynth: Leveraging Large Language Models for Circuit Topology Generation}}
\end{abstract}

\section{Introduction}

Circuit topology synthesis stands as a complex and critical aspect of electronic circuit design. The configuration and interconnection of components directly influence critical circuit functionality and performance. With the increasing demands for integration and complexity in modern electronic systems, the role of circuit topology synthesis becomes crucial in meeting design specifications and performance criteria. However, relying solely on human intervention for topology synthesis is a formidable challenge. As the complexity of contemporary circuit designs grows, the search space expands exponentially, rendering exhaustive or random exploration impractical.

Traditional methods, encompassing rule-based systems, heuristic approaches, and genetic algorithms, have been proposed in the past to tackle circuit topology synthesis. Yet, these methods often encounter limitations in scalability, adaptability to evolving design requirements, and efficiency. While there are different approaches proposed for circuit design process, using Large Language Models (LLMs) for circuit topology synthesis remains relatively underexplored. While LLMs have showcased exceptional capabilities in natural language understanding and generation \cite{zhao2023survey}, their application in circuit synthesis remains largely untapped. The ability of LLMs to learn complex patterns, comprehend relationships, and generate diverse outputs holds promise for overcoming the limitations of traditional methods.

In this context, we introduce \textsc{CircuitSynth}, an innovative approach that harnesses the power of LLMs for automated circuit topology synthesis. Figure \ref{fig:arch} provides an overview of the proposed method, with the goal of generating a valid circuit topology in the form of a netlist based on a text prompt. \textsc{CircuitSynth} employs a sophisticated methodology, leveraging an extensive dataset of valid and invalid circuit configurations. 

% Our approach has four key steps. In the first step, we curated a dataset of valid and invalid power converter circuit designs with 5 device components using Random Search (RS) \cite{fan2021specification} and NGSpice \cite{nenzi2011ngspice} simulator.
% Next, we train a circuit validity classifier that takes the circuit represented as a netlist as input and produces a probability of a circuit being valid. Subsequently, we fine-tune an LLM using the valid circuits in the dataset for circuit topology generation. Finally, we fine-tune the LLM with a loss

% Initially, we train a circuit validity classifier that takes the circuit represented as a netlist as input and produces a probability of a circuit being valid. Subsequently, we fine-tune the LLM using the valid circuits in the dataset for circuit topology generation. Finally, we refine model outputs by scoring them using the circuit validity classifier and updating them to minimize the circuit invalidity. During this refinement step, the non-differentiability of the LLM outputs due to discrete choices presents a challenge for gradient-based optimization. The Gumbel-Max trick offers a solution by introducing continuous relaxation to the discrete choices, ensuring effective gradient-based optimization and improving the overall validity of the circuit generated by the LLM. 

Our proposed approach adopts a two-phase model architecture, comprising Circuit Topology Generation and Circuit Topology Refinement. In the initial phase, we fine-tune a LLM to produce a circuit topology in an autoregressive manner. % probability distribution over tokens at each time step. This distribution is then used to sample the circuit topology in an autoregressive manner. 
Since the generated circuits may not consistently meet validity constraints, in the Circuit Topology Refinement phase we undertake two pivotal steps. Firstly, we employ a classifier to gauge the likelihood that a generated circuit topology adheres to the requisites of a valid circuit design. This ensures alignment with the necessary design parameters. Secondly, in Generation Enhancement we refine the circuit topology generation process, by minimizing the combined loss of negative log-likelihood with the circuit invalidity score. % Throughout this stage, the circuit validity classifier remains unchanged to uphold efficient training. It should be noted that during the refinement phase, the non-differentiability of drawing discrete samples from the LLM output distribution poses a challenge for gradient-based optimization. To address this, we employ the Gumbel-Max trick, which introduces continuous relaxation to the discrete choices. This facilitates effective gradient-based optimization and enhances the overall validity of the circuits generated by the LLM. 
We evaluate our model by generating new circuit topologies using the trained model and passing them to the SPICE simulator to verify their validity. Experimental results demonstrate the effectiveness of \textsc{CircuitSynth} compared to various LLM variants, emphasizing its potential for automating circuit topology synthesis. 

\textbf{Contributions:} The key contributions are summarized as follows: (a) Introduction of \textsc{CircuitSynth}, a novel methodology leveraging LLMs for automated circuit topology synthesis; (b) Generation of a comprehensive dataset encompassing both valid and invalid circuit configurations, facilitating the development and evaluation of \textsc{CircuitSynth}; (c) Development of a circuit validity classifier to assess a circuit's validity, enhancing the reliability of \textsc{CircuitSynth} outputs.

\section{Our Approach}
\subsection{Dataset}
We curated a dataset of power converter circuit designs with 5 device components using Random Search (RS) \cite{fan2021specification} and NGSpice \cite{nenzi2011ngspice} simulator. Together we collected a total of 862,606 circuits (567,307 valid and 295,299 invalid ones). In this study we only consider devices including capacitors $C$, inductors $L$, phase-I switches $S_a$ and phase-II switches $S_b$. More details about our dataset are described in Appendix \ref{app:dataset}.

\begin{figure}[!ht]
    \centering
    \includegraphics[width=0.35\textwidth]{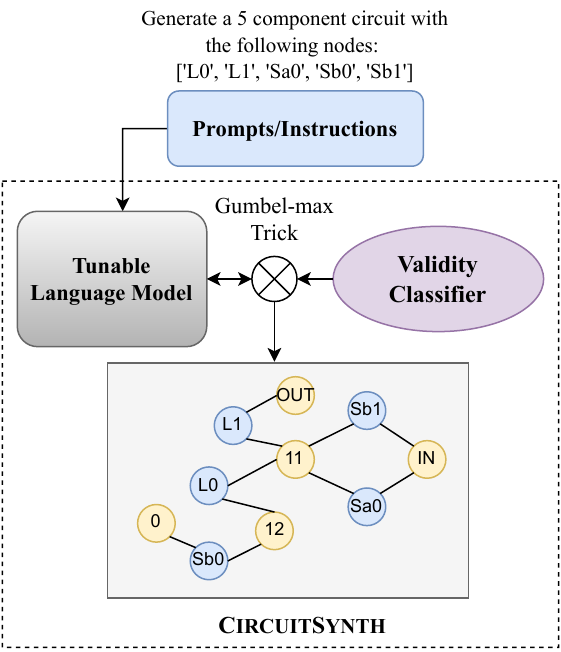}
    \caption{An overview of our proposed \textsc{CircuitSynth}. Given a natural language prompt with the component pool, our model can explore the design space effectively and generate circuit topologies by leveraging LLMs.}
    \label{fig:arch}
\end{figure}

\subsection{Model Overview}
Figure \ref{fig:arch} illustrates the model architecture, comprising two main phases: Circuit Topology Generation and Circuit Topology Refinement. In the first phase, we utilize a large language model (LLM) to generate a probability distribution $p^t_{LLM}$ over tokens at each time step $t$, given an input prompt $X=[x^{(1)}, \ldots ,x^{(L)}]$ of sequence length $L$ containing a component pool. From this distribution, we sample the circuit topology $Y = [y^{(1)}, \ldots, y^{(t)}, \ldots ,y^{(N)}]$ in an autoregressive manner, where $N$ denotes the sequence length. It should be noted that the generated circuit may not always adhere to validity constraints. To address this, the second phase, Circuit Topology Refinement, consists of two steps:

\subsubsection{Circuit Validity Estimation} Here, we implement a classifier that assesses the probability $p_{\text{valid}}$ that the generated circuit topology, represented as a netlist, is valid. This classifier evaluates the compliance of the generated topology with the requirements of a valid circuit design.    
\subsubsection{Generation Enhancement} %Upon identifying potential invalidities, the model aims to refine the circuit topology generation process. 
This phase enhances the LLM for valid circuit generation by minimizing the combined loss of negative log-likelihood with the circuit invalidity score. During this stage, the circuit validity classifier remains frozen to ensure efficient training.

Estimating circuit validity involves drawing discrete samples using a categorical distribution from the language model, which poses challenges for gradient-based optimization. To overcome this, we employ the Gumbel softmax trick \cite{maddison2016concrete,jang2016categorical}. Gumbel-Softmax relaxation between the topology generator and the circuit validity classifier enables the backpropagation of gradients among discrete circuit topology samples during training. We describe these methods in details in the subsequent sections.

\subsection{Circuit Topology Generation}

We formulate the task of circuit topology generation as a text generation problem. In this paper, we work with datasets $\mathcal{D} = \{(X_i, Y_i)\}_{i=1}^{T}$ where $X_i$ refers to the input prompt and $Y_i$ refers to the valid circuit topology netlist. Each entry in the netlist corresponds to a node in an undirected graph $\mathcal{G}$, with edges denoting interconnections between these nodes. The choice of encoding strategy for representing the netlist textually plays a pivotal role, as it significantly influences the effectiveness of LLMs in our task. Therefore, we adopt the ``Incident" encoding strategy, known for its superiority over other encoding methods in various graph-related tasks \cite{fatemi2024talk}. Figure \ref{fig:incident} in Appendix \ref{app:incident_enc} demonstrates an example of how a netlist is represented using the incident encoding method.

Formally, the circuit topology generation process can be described as:
\begin{equation}
Y \sim \mathbb{E}_{X\sim\mathcal{D}}[p_{\text{LLM}}(Y|X)]
\end{equation}
Where $p_{\text{LLM}}$ is the pretrained LLM parameterized by a set of parameters $\theta$. This can be further decomposed into:
\begin{equation}
p_{\text{LLM}}(Y|X) = \prod_{t=1}^{N} p_{\text{LLM}}(y_t|X,\{y_j\}_{j=1}^{t-1})
\end{equation}
To tune the pretrained LLM, we use the conventional negative log-likelihood objective:
\begin{equation}
\mathcal{L}_{\text{LLM}} = \sum_{t=1}^{N} -\log p_{\text{LLM}}(y_t|X,\{y_j\}_{j=1}^{t-1})
\end{equation}
This objective generally ensures fluency of text in the incident encoding method. However, this objective can be incomplete. The circuit topology generated may not include all the components in the component pool or satisfy circuit validity constraints. In order to ensure the circuits meet additional constraints, we introduce the circuit topology refinement phase that learns to incorporate specific constraints into the circuit topology generation process.

\subsection{Circuit Topology Refinement}

%The purpose of the circuit topology refinement phase is to ensure that specific circuit topology design constraints are met during the circuit topology generation process. 
This phase comprises two main steps: (a) Circuit Validity Estimation, which evaluates whether the generated circuit topology adheres to necessary constraints, and (b) Generation Enhancement, which utilizes feedback from the constraint estimation to improve the overall circuit topology generation process. In this study, the primary focus is on optimizing for circuit validity. We elaborate on these steps below.

\subsubsection{Circuit Validity Estimation}

%Circuit validity estimation involves determining the probability of a circuit topology being valid. For this purpose, 
We train a dedicated classifier, $f_{\text{valid}}$, on the dataset $\mathcal{D} = \{\mathcal{D}_{\text{valid}} \cup \mathcal{D}_{\text{invalid}}\}$, which includes both valid and invalid circuit topologies. The output of the classifier, $p_{\text{valid}}$, represents the probability of validity for a given circuit topology. We employ a RoBERTa-based classifier \cite{liu2019roberta} optimized with binary cross-entropy loss, achieving an $F_1$ score of 92\% for binary classification of circuit validity.

\subsubsection{Generation Enhancement}

To refine the circuit topology generation process, we utilize the pretrained circuit validity classifier to compute a circuit validity loss:
\begin{equation}
\mathcal{L}_{\text{\text{valid}}} = (1 - p_{\text{\text{valid}}})
\end{equation}
Subsequently, we combine this circuit validity loss with the standard negative log-likelihood loss $\mathcal{L}_{\text{LLM}}$ to ensure adherence to circuit validity constraints. Formally, the combined loss function is defined as:
\begin{equation}
\mathcal{L} = \lambda_1 \mathcal{L}_{\text{LLM}} + \lambda_2 \mathcal{L}_{\text{valid}}
\label{eq:loss}
\end{equation}
Here, $\lambda_1$ and $\lambda_2$ are learnable coefficients that determine the relative importance of each loss component.

During training, a significant challenge arises when sampling $\hat{Y}$ from the distribution $p_{\text{LLM}}$ to feed into the circuit validity classifier. This sampling process involves drawing samples from a non-differentiable categorical distribution, hindering gradient propagation \cite{nie2018relgan}. To address this issue, we employ the Gumbel-softmax relaxation \cite{maddison2016concrete,jang2016categorical} to approximate the discrete sampling process $\hat{y}^{(t)} \sim p_{\text{LLM}}$. Firstly, we apply the Gumbel-Max trick to reparameterize sampling from $p_{\text{LLM}}$. This is given as follows:
\begin{gather}
u^{(i)} \sim \text{uniform}(0, 1) \\
z^{(i,t)} = -\log(-\log(u^{(i)})) \\
\hat{y}^{(t)} = \text{one-hot}\left[\text{argmax}_{i\in|V|}(\hat{p}^{(i,t)}_{\text{LLM}} + z^{(i,t)})\right]
\label{eq:gumbel}
\end{gather}

Where $|V|$ is the size of the vocabulary, $\hat{p}^{(i,t)}_{\text{LLM}}$ refers to the logits, i.e., pre-softmax activation of $p_{\text{LLM}}$ at the $t$-th generation step for the $i$-th word, and $z^{(i,t)}$ are i.i.d. samples from the standard Gumbel distribution. 
Next, we approximate the discrete $argmax$ operation in Equation \ref{eq:gumbel} with the continuous softmax operator to ensure differentiability as:
\begin{equation}
\tilde{y}^{(t)} = \text{softmax}\left(\frac{\hat{p}^{(i,t)}_{\text{LLM}} + z^{(i,t)}}{\tau}\right)
\end{equation}
where $\tau$ is a temperature hyperparameter, which controls how close $\tilde{y}^{(t)}$ is to $\hat{y}^{(t)}$. %$\tilde{y}^{(t)} = \hat{y}^{(t)}$ when $\tau = 0$. 
Finally, to enable gradient flow during training, we utilize the straight-through estimator \cite{bengio2013estimating}. In this approach, we use $\hat{y}^{(t)}$ in the forward pass and $\tilde{y}^{(t)}$ in the backward pass, allowing for efficient backpropagation.

\section{Experiments}
In this section, we provide details about the models, baselines and metrics used to train and evaluate. We trained and compared two LLMs for \textsc{CircuitSynth}: GPT-Neo-2.7 and StableLM-3B-4E1T (refereed to as GPT-Neo and StableLM in subsequent sections). More model descriptions and implementation details are provided in Appendix \ref{app:models} and Appendix \ref{app:implement}, respectively.

\subsection{Baselines}
We conducted experiments with the following baselines:% and model variants:
\subsubsection{Zero-Shot Generation} We provide prompts containing the components pool as input to LLMs including Llama-2 (13b) \cite{touvron2023llama} and Flan-Ul2 (20b) \cite{tay2022ul2} without any fine-tuning.% and attempt to generate the circuit topology directly.
    
\subsubsection{In-Context Learning (ICL)} %Leveraging the in-context learning ability of LLMs such as Llama-2 (13b) and Flan-Ul2 (20b), 
We provide demonstrations of input prompts with component pools and example circuits preceding a new input prompt to Llama-2 (13b) and Flan-Ul2 (20b) models. We limit the number of in-context examples to $k \in \{5, 10, 20\}$ and experiment with incident encoding and netlist array-like structures for representing circuit topologies.
    
\subsubsection{Parameter-Efficient Fine-Tuning (PEFT)} We adopt simple PEFT-based approaches \cite{houlsby2019parameter} to tune Llama-2 (13b) and Flan-Ul2 (20b) models to investigate if a relatively smaller language model (GPT-Neo/StableLM) trained using our approach can achieve comparable performance to much larger LLMs PEFT-tuned for the same task. 
We performed Prompt-tuning which involves learning task-specific soft prompts and adding them to the input while keeping the pre-trained model parameters frozen. In our study, we explore different numbers of trainable soft prompt tokens, ranging from 100 to 500.

\subsubsection{Vanilla Fine-Tuning} GPT-Neo and StableLM models are fine-tuned with the objective of minimizing the negative log-likelihood for generating circuit topologies.
    
\subsubsection{\textsc{CircuitSynth} (CS)} We introduce \textsc{CircuitSynth}, our complete framework aimed at enhancing fine-tuned circuit topology generation using a circuit validity classifier to improve the validity of generated circuit topologies.

\subsection{Metrics}
In our evaluation setup, we report the fraction of unique circuit topologies that are estimated as valid %as $E(f_{clf}(\hat{y}))$, where $clf \in \{\text{valid}, \mathcal{S}_{\text{valid}}\}$ refers to the validity estimated 
by the independent validity classifier and the SPICE simulator as $E(f_{\text{valid}}(\hat{y}))$ and $E(f_{\mathcal{S}_{\text{valid}}}(\hat{y}))$ respectively, based on 1000 samples of unique circuit topologies. %$f_{\text{valid}}(\hat{y})$ uses the independent classifier to estimate the validity of the generated circuit topologies, while $f_{\mathcal{S}_{\text{valid}}}(\hat{y})$ refers to the SPICE Simulator results that ascertain the validity. 
 In our experiments, a circuit topology is considered valid if it has a validity score, surpassing 0.6, from the circuit validity classifier. Additionally, we report the efficiency of the generated circuit as computed using the SPICE simulator as $E(f_{\mathcal{S}_{\text{eff}}}(\hat{y}))$. The SPICE simulator evaluates the validity and efficiency of a given netlist by verifying its electrical characteristics and performance metrics through detailed circuit simulations. It evaluates parameters such as voltage levels, timing, and power consumption of the circuit topology with certain duty cycles under given conditions. Furthermore, we calculate a Duplicate Generation Rate (DGR), denoted by $\rho$, defined as ``\text{\# topologies generated}'' divided by ``\text{\# unique topologies}''.% which indicates the number of circuit topologies required to be sampled from the model to obtain a unique circuit topology design.  $\rho$ is defined as ``\text{\# circuit topologies generated}'' divided by ``\text{\# unique circuit topologies}''.
% Formally,
% \begin{equation}
% \rho = \frac{\text{\# circuit topologies generated}}{\text{\# unique circuit topologies}}
% \end{equation}

% \usepackage{booktabs}
\begin{table}[h!]
\small
\centering
\caption{Evaluation Results. Llama and Flan refers to Llama-2 (13b) and Flan-Ul2 (20b) models respectively. $\rho$ denotes the DGR metric. CS refers to our \textsc{CircuitSynth} model. \label{tab:eval_res}}
\begin{tabular}{@{}lcccc@{}}
\toprule
\multicolumn{1}{l|}{\textbf{Models}} &
  \multicolumn{1}{c|}{$E(f_{\text{valid}}(\hat{y}))$} &
  \multicolumn{1}{l|}{$E(f_{\mathcal{S}_{\text{valid}}}(\hat{y}))$} &
  \multicolumn{1}{c|}{$E(f_{\mathcal{S}_{\text{eff}}}(\hat{y}))$} &
  $\rho$ \\ \midrule
\multicolumn{5}{c}{\textbf{PEFT Generation}} \\ \midrule
\multicolumn{1}{l|}{Llama$_{p_{500}}$} &
  \multicolumn{1}{c|}{0.526} &
  \multicolumn{1}{c|}{0.337} &
  \multicolumn{1}{c|}{0.611} &
  4.45 \\
\multicolumn{1}{l|}{Llama$_{p_{100}}$} &
  \multicolumn{1}{c|}{0.581} &
  \multicolumn{1}{c|}{0.648} &
  \multicolumn{1}{c|}{0.576} &
  3.93 \\
\multicolumn{1}{l|}{Flan$_{p_{100}}$} &
  \multicolumn{1}{c|}{\textbf{0.608}} &
  \multicolumn{1}{c|}{\textbf{0.663}} &
  \multicolumn{1}{c|}{\textbf{0.694}} &
  \textbf{3.65} \\ \midrule
\multicolumn{5}{c}{\textbf{Fine-tuned Generation}} \\ \midrule
\multicolumn{1}{l|}{GPT-Neo$_{ft}$} &
  \multicolumn{1}{c|}{0.591} &
  \multicolumn{1}{c|}{0.60} &
  \multicolumn{1}{c|}{0.692} &
  1.89 \\
\multicolumn{1}{l|}{StableLM$_{ft}$} &
  \multicolumn{1}{c|}{0.598} &
  \multicolumn{1}{c|}{0.591} &
  \multicolumn{1}{c|}{0.682} &
  1.85 \\
\multicolumn{1}{l|}{\textsc{CS}$_{GPT-Neo}$} &
  \multicolumn{1}{c|}{\textbf{0.636}} &
  \multicolumn{1}{c|}{\textbf{0.648}} &
  \multicolumn{1}{c|}{0.713} &
  1.31 \\
\multicolumn{1}{l|}{\textsc{CS}$_{StableLM}$} &
  \multicolumn{1}{c|}{0.632} &
  \multicolumn{1}{c|}{0.624} &
  \multicolumn{1}{c|}{\textbf{0.728}} &
  \textbf{1.31} \\ \bottomrule
\end{tabular}
\end{table}

% an independent circuit validity classifier, $f_{valid}$, and the SPICE simulator to assess performance. The classifier demonstrates a high accuracy rate exceeding 92\%. At the end of each epoch of the training phase, we generate around 100 unique circuit topologies and utilize this classifier to evaluate them for validity. Subsequently, we juxtapose these classifier results with those obtained from the SPICE simulator, which serves as a benchmark for validating the circuit topologies generated by the models employed in this study.  \label{tab:res}

% \subsection{Implementation Details}

\section{Results \& Discussion}

\subsection{Overview}
Table \ref{tab:eval_res} presents the evaluation results of our circuit synthesis models, showing the ratios of valid, unique, and efficient circuits, along with the DGR ($\rho$). Each model is assessed by generating 1000 unique sample circuit topologies. Our findings suggest that \textsc{CircuitSynth} models consistently outperforms most baselines across various evaluation metrics. Note that the smaller language models tuned using our method perform comparably with larger prompt-tuned models. 

\subsection{Performance of Zero-Shot and ICL Methods}

We observe a significant performance gap between zero-shot generation methods and fine-tuned approaches. Zero-shot generation struggles to produce valid netlist-like structures for subsequent classification or simulation. Even with ICL, where the model is provided with examples, %demonstrations of sample prompts and corresponding circuit generations, 
there is only a marginal improvement and the completion rate for all components remains unsatisfactory. Furthermore, increasing the number of in-context examples $k$ resulted in diminishing returns. 

\subsection{Effectiveness of \textsc{CircuitSynth}}

\subsubsection{Comparison with Similar-sized Model Fine-tuning}

Our \textsc{CircuitSynth} models demonstrate a marked superiority in terms of generating valid circuits compared to vanilla fine-tuned GPT-Neo and StableLM architectures. Specifically, the \textsc{CircuitSynth} model utilizing GPT-Neo demonstrates a notable enhancement in circuit validity and efficiency compared to its vanilla counterpart. %We emphasize that our comparison is conducted on models of similar sizes, highlighting that despite this similarity, our method consistently outperforms vanilla fine-tuned models. 
As both models have a similar size, this observation underscores the efficacy of our approach in improving the synthesis of valid circuit topologies. Also, the duplicate generation rate for our \textsc{CircuitSynth} models is significantly lower compared to other fine-tuned models, indicating faster production of unique circuit topologies.

\subsubsection{Comparison with PEFT-tuned Models}

Our experiments demonstrate that PEFT-based prompt tuning of a Flan-ul2 model ($\sim20$-billion parameters), yields performances comparable to that of our models with $\sim{3}$-billion parameters. Notably, we find that a smaller language model (GPT-Neo/StableLM) trained using our approach can achieve comparable performance to LLMs fine-tuned for the same task. The duplicate generation rate of our approach is much lower than the prompt-tuned LLama-2/Flan-ul2 models, which indicates that our method empowers a smaller language model to produce unique circuit topologies faster than PEFT-tuned 20 billion parameter models. An intriguing finding is the capacity of a smaller fine-tuned model to effectively capture extensive information from a larger prompt-tuned model. This highlights the effectiveness of our proposed methodology.

\subsection{Validity Correlation}
To evaluate the correlation between classifier prediction probabilities and simulator-assessed ground truth validity scores, we used a two-sample t-test. We analyzed 1000 sample circuit topology generations each from \textsc{CS}$_{GPT-Neo}$ and \textsc{CS}$_{StableLM}$. The classifier prediction probabilities were treated as the continuous predictor variable, while the binary ground truth validity served as the outcome variable. With a statistically significant $p<0.05$, we found a strong correlation between the classifier predictions and the ground truth validity scores provided by the simulator.

%The t-test results affirm the reliability and accuracy of the classifier in capturing the underlying patterns of the ground truth validity, underscoring its effectiveness in the given application.

% Additionally, we compute correlation scores between our classifier-based validity predictions and simulator-based validity. With a very strong Pearson correlation score of 0.843, we find that our classifier serves as a good proxy for the simulator during training. More results and discussion are provided in Appendix \ref{app:validity_corr}.

\subsection{Ablation Study}

%We further present an ablation study aimed at investigating the impact of natural language incident encoding on tuning the model's performance. For \textsc{CircuitSynth}$_{GPT-Neo}$, we observe a significant improvement in both metrics when employing natural language incident encoding. For \textsc{CircuitSynth}$_{StableLM}$, the impact of natural language incident encoding is less pronounced. Detailed results and discussion about this part is provided in Appendix \ref{app:ablation}.

We present an ablation study to investigate the impact of natural language incident encoding on the model's performance by replacing the encoding with an array-like structure to represent netlists (see the left part of Figure \ref{fig:incident}). 
Table \ref{tab:ablation} shows the results of our evaluation of \textsc{CS}$_{GPT-Neo}$ and \textsc{CS}$_{StableLM}$ with and without natural language incident encoding. For \textsc{CS}$_{GPT-Neo}$, we observe significant improvement in both metrics with natural language incident encoding. Conversely, for \textsc{CS}$_{StableLM}$, the impact is less pronounced. We hypothesize that this disparity is due to differences in their training data. StableLM, trained on both natural language and coding datasets, may rely less on explicit natural language representations for effective circuit synthesis. Thus, our ablation study shows that the benefits of natural language incident encoding depend on the model architecture.

% As illustrated in Table \ref{tab:ablation}, \textsc{CircuitSynth}$_{GPT-Neo}$ and \textsc{CircuitSynth}$_{StableLM}$, are evaluated with and without natural language incident encoding. For \textsc{CircuitSynth}$_{GPT-Neo}$, we observe a significant improvement in both metrics when employing natural language incident encoding. Conversely, for \textsc{CircuitSynth}$_{StableLM}$, the impact of natural language incident encoding is less pronounced. We hypothesize that the disparity in the effectiveness of natural language incident encoding between GPT-Neo and StableLM could be attributed to the difference in their training data. StableLM is trained on a diverse dataset that includes both natural language text and coding datasets, potentially reducing the reliance on explicit natural language representations for effective circuit synthesis. Therefore, our ablation study highlights that the benefits of natural language incident encoding depend on the underlying model architecture. 

\begin{table}[tb!]
\centering
\caption{Ablation Study: Effect of natural language (NL) incident encoding for tuning the model.\label{tab:ablation}}
\begin{tabular}{@{}lll@{}}
\toprule
\textbf{Models} & $E(f_{\mathcal{S}_{\text{valid}}}(\hat{y}))$ & $E(f_{\mathcal{S}_{\text{eff}}}(\hat{y}))$ \\ \midrule
\textsc{CircuitSynth}$_{GPT-Neo}$  & \textbf{0.648 }& \textbf{0.713} \\
\quad w/o NL                      & 0.628 & 0.677 \\ \midrule
\textsc{CircuitSynth}$_{StableLM}$ & 0.624 & \textbf{0.728 }\\
\quad w/o NL                      & \textbf{0.640}  & 0.685 \\ \bottomrule
\end{tabular}
\end{table}

\subsection{Potential Emergent Capability: Efficiency Scores}

An emergent capability of our models is the production of circuit topologies with good efficiency scores. Despite the primary objective being the generation of valid circuits, our models demonstrate the additional ability to optimize for efficiency, further enhancing their practical utility.

\section{Conclusion}
In this study, we introduced \textsc{CircuitSynth}, a novel method that harnesses LLMs to automate circuit topology synthesis. Leveraging a circuit validity classfier and the Gumbel-softmax trick, \textsc{CircuitSynth} was trained to enhance the overall validity of generated circuits. Experimental results demonstrate that \textsc{CircuitSynth} yields significant improvements across multiple metrics when compared to various LLM variants. \textsc{CircuitSynth} offers a promising avenue for revolutionizing electronic circuit design, promising advancements in efficiency, performance, and scalability.

\bibliographystyle{IEEEtran}
\bibliography{conf}
% \bibitem{b1} G. Eason, B. Noble, and I. N. Sneddon, ``On certain integrals of Lipschitz-Hankel type involving products of Bessel functions,'' Phil. Trans. Roy. Soc. London, vol. A247, pp. 529--551, April 1955.
% \bibitem{b2} J. Clerk Maxwell, A Treatise on Electricity and Magnetism, 3rd ed., vol. 2. Oxford: Clarendon, 1892, pp.68--73.
% \bibitem{b3} I. S. Jacobs and C. P. Bean, ``Fine particles, thin films and exchange anisotropy,'' in Magnetism, vol. III, G. T. Rado and H. Suhl, Eds. New York: Academic, 1963, pp. 271--350.
% \bibitem{b4} K. Elissa, ``Title of paper if known,'' unpublished.
% \bibitem{b5} R. Nicole, ``Title of paper with only first word capitalized,'' J. Name Stand. Abbrev., in press.
% \bibitem{b6} Y. Yorozu, M. Hirano, K. Oka, and Y. Tagawa, ``Electron spectroscopy studies on magneto-optical media and plastic substrate interface,'' IEEE Transl. J. Magn. Japan, vol. 2, pp. 740--741, August 1987 [Digests 9th Annual Conf. Magnetics Japan, p. 301, 1982].
% \bibitem{b7} M. Young, The Technical Writer's Handbook. Mill Valley, CA: University Science, 1989.
% \end{thebibliography}

\appendices
\setcounter{figure}{0}
\renewcommand\thefigure{\Alph{section}.\arabic{figure}}

\section{Dataset}\label{app:dataset}
We curated a dataset of power converter circuit designs with 5 device components. We first used Random Search (RS) \cite{fan2021specification} to generate numerous random circuit topologies. Then we utilized an open-source electronic circuit simulator NGSpice \cite{nenzi2011ngspice} to identify valid and invalid circuits and collected efficiency values for the valid circuits. Together we collected a total of 862,606 circuits, with 567,307 valid ones and 295,299 invalid ones. In this study we only consider devices including capacitors $C$, inductors $L$, phase-I switches $S_a$ and phase-II switches $S_b$. Each device component has two ports. Together with the three external ports $Vin$, $Vout$ and $Gnd$ (renamed with 'IN', 'OUT' and '0', respectively), the design space contains topologies with 13 ports in total. The device ports are indexed by numbers and the connected ports are represented by only one of the port numbers (see example in Figure. \ref{fig:arch}). We used fixed device parameters for capacitors (10$\mu F$) and inductors (100$\mu H$). For external ports, we use an input resistor of 0.1$\Omega$ for $Vin$, and an output resistor of 50$\Omega$ and an output capacitor of 10$\mu F$ for $Vout$. The duty cycle is randomly selected from a set of [0.1, 0.3, 0.5, 0.7, 0.9]. The frequency and input voltage are configured as 1$MHz$ and 100$V$, respectively.

\section{Incident Encoding: Example}
\label{app:incident_enc}
\begin{figure}[h!]
    \centering
    \includegraphics[width=0.35\textwidth]{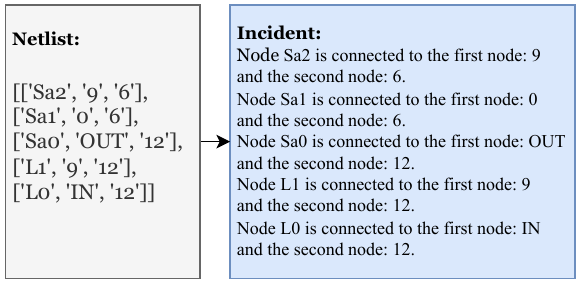}
    \caption{Example: incident encoding of a netlist.}
    \label{fig:incident}
\end{figure}

%\section{Incident Encoding}\label{app:incident_enc}
%Figure \ref{fig:incident} shows an example of how a netlist is represented using the incident encoding method.

%\begin{figure}[h!]
%    \centering
%    \includegraphics[width=0.45\textwidth]{incident.pdf}
%    \caption{Example: incident encoding of a netlist.}
%    \label{fig:incident}
%\end{figure}

\section{LLM Models}\label{app:models}
\label{sec:models}
More detailed descriptions of the LLMs used for our circuit topology generation task are provided below:  
\begin{itemize}
    \item GPT-Neo-2.7\cite{black2021gpt} \footnote{{https://huggingface.co/EleutherAI/gpt-neo-2.7B}}: GPT-Neo 2.7B is a transformer model designed using EleutherAI's replication of the GPT-3 architecture. GPT-Neo refers to the class of models, while 2.7B represents the number of parameters of this particular pre-trained model.
    
    \item StableLM-3B-4E1T \footnote{{https://huggingface.co/stabilityai/stablelm-3b-4e1t}}: StableLM-3B-4E1T is a 3 billion parameter decoder-only language model pre-trained on 1 trillion tokens of diverse English and code datasets for 4 epochs. We refer to this as StableLM in subsequent sections.
\end{itemize}

\section{Implementation Details}\label{app:implement}
We provide the implementation details of our experiments conducted with the official PyTorch v2.2.0 release binary package, compiled with CUDA 11.8, utilizing NVIDIA V100 GPUs with 32 GB of memory.

\begin{itemize}
\item Training: We utilize shuffled data from the training split for 5-7 epochs, saving the model checkpoint with the best performance on the validation split. To conserve memory, we implement gradient checkpointing. Additionally, we employ the AdamW optimizer \cite{loshchilov2017decoupled} with beta parameters set to 0.9 and 0.95 respectively, and an epsilon value of 1.0E-8. Moreover, we set the learning rate to 0.95E-5 and fix the seed to 42 for reproducibility purposes.
\item Inference: We assess all models by generating 1000 unique sample circuit topologies using each model. These instances are generated through a combination of nucleus sampling and top-k sampling techniques. Subsequently, the generated circuit topologies are inputted into the validity classifier to determine the percentage of valid generations. During training, we evaluate a subset of 100 sample generations and employ identical evaluation settings to monitor performance, saving the checkpoint if the current performance surpasses that of the previous epoch.
\end{itemize}

% \section{Validity Correlation}
% \label{app:validity_corr}
% \begin{figure}[!ht]
%     \centering
%     \includegraphics[width=0.48\textwidth]{correlation_plt.pdf}
%     \caption{Pearson Correlation between Classifier-predicted validity and simulator-based validity.}
%     \label{fig:corr_plt}
% \end{figure}

% Figure \ref{fig:corr_plt} presents the Pearson correlation plot between the validity classifier predictions and the validity determined by the SPICE simulator. The plot indicates a very strong correlation. However, some noisy predictions from the classifier could lead to performance degradation. Despite this, our approach effectively enhances the overall validity of the generated circuit topologies. In future work, we will explore further improvements and the impact of classifiers with varying validity performance.

\end{document}